\documentclass[runningheads]{llncs}
\usepackage{xcolor}
\usepackage{subcaption}
\usepackage{hyperref}
\usepackage{graphicx}
\usepackage[export]{adjustbox}
\usepackage{float}
\usepackage{chngcntr}

\begin{document}
\title{Surface-based parcellation and vertex-wise analysis of ultra high-resolution \textit{ex vivo} 7 tesla MRI in Alzheimer's disease and related dementias}
\titlerunning{7T ex vivo brain MRI analysis}

\author{Pulkit Khandelwal\inst{1},
Michael Tran Duong\inst{1},
Lisa Levorse \inst{1},
Constanza Fuentes\inst{2},
Amanda Denning\inst{1},
Winifred Trotman\inst{1},
Ranjit Ittyerah\inst{1},
Alejandra Bahena\inst{1},
Theresa Schuck\inst{1},
Marianna Gabrielyan\inst{1},
Karthik Prabhakaran\inst{1},
Daniel Ohm\inst{1},
Gabor Mizsei\inst{1},
John Robinson\inst{1},
Monica Muñoz\inst{2},
John Detre\inst{1},
Edward Lee\inst{1},
David Irwin\inst{1},
Corey McMillan\inst{1},
M. Dylan Tisdall\inst{1},
Sandhitsu Das\inst{1},
David Wolk\inst{1},
\and Paul A. Yushkevich\inst{1}}
\authorrunning{Khandelwal et al.}

\institute{University of Pennsylvania, Philadelphia, USA \and
University of Castilla-La Mancha, Ciudad Real, Spain\\
\email{pulks@seas.upenn.edu}}
\maketitle

\begin{abstract}
Magnetic resonance imaging (MRI) is the standard modality to understand human brain structure and function \textbf{\textit{in vivo}} (antemortem). Decades of research in human neuroimaging has led to the widespread development of methods and tools to provide automated volume-based segmentations and surface-based parcellations which help localize brain functions to specialized anatomical regions. Recently \textbf{\textit{ex vivo}} (postmortem) imaging of the brain has opened-up avenues to study brain structure at sub-millimeter ultra high-resolution revealing details not possible to observe with \textit{in vivo} MRI. Unfortunately, there has been limited methodological development in \textit{ex vivo} MRI primarily due to lack of datasets and limited centers with such imaging resources. Therefore, in this work, we present one-of-its-kind dataset of \textbf{82 \textit{ex vivo} T2w whole-brain hemispheres MRI at 0.3 mm$^3$} resolution spanning Alzheimer's disease and related dementias. We adapted and \textbf{developed a fast and easy-to-use} automated surface-based pipeline to parcellate, for the first time, ultra high-resolution \textit{ex vivo} brain tissue at the native subject-space resolution using the Desikan-Killiany-Tourville (DKT) brain atlas. This allows us to perform vertex-wise analysis in the template space and thereby link morphometry measures with pathology measurements derived from histology. We open-source our dataset, docker container, Jupyter notebooks for ready-to-use out-of-the-box set of tools and command line options to advance \textit{ex vivo} MRI clinical brain imaging research at the \href{https://pulkit-khandelwal.github.io/exvivo-brain-upenn/}{\textbf{project webpage}}.
\end{abstract}

\section{Introduction}
\label{sec:Introduction}
\vspace{-3mm}
\hspace{5 mm} Alzheimer's disease and related dementias (ADRD) are heterogeneous, with multiple neuropathological processes jointly contributing to neurodegeneration \cite{robinson}. Currently, co-pathologies are not reliably detected with \textit{in vivo} imaging, which makes it difficult for clinicians to accurately diagnose the cause of cognitive impairment in individual patients and to identify those most likely to benefit from treatment with emergent drugs that target AD pathology. The efficacy of \textit{in vivo} biomarkers can be enhanced by combined analysis of regional measures of neurodegeneration, such as cortical thickness, derived from brain MRI (\textit{ex vivo} or \textit{in vivo}) and pathological measures obtained at autopsy. Such analysis can help identify distinct spatial patterns of disease progression linked to AD-specific pathologies (i.e. $\beta$-amyloid plaques and phosphorylated tau protein tangles) and common co-pathologies. While such studies are performed using either \textit{in vivo} or \textit{ex vivo} MRI, the latter offers advantages in temporal proximity to pathological markers and in spatial resolution. The ability to scan \textit{ex vivo} brain tissue for many hours without motion artifacts makes it possible to achieve ultra-high spatial resolution at which intricate neuroanatomical details of the brain can be revealed [\cite{augustinack}, \cite{Iglesias_hippo}, \cite{alkemade}, \cite{adler}, \cite{Yushkevich_brain} , \cite{sadhana_acta}, \cite{Tisdall}, and \cite{deKraker}]. \textbf{However, lack of automated tools for processing ultra high-resolution \textit{ex vivo} MRI is a significant barrier to study structure-pathology association in ADRD}.

Decades of neuroimaging research has yielded advanced computational frameworks for automated analysis of \textit{in vivo} brain MRI, and tools such as  FreeSurfer \cite{Fischl_one} and Statistical Parametric Mapping (SPM) \cite{SPM}, FSL \cite{FSL} have been applied in a plethora of ADRD neuroimaging studies. However, these tools cannot be directly used on ultra-high resolution \textit{ex vivo} MRI and  there is very limited work on developing \textit{ex vivo} MRI analysis methods for widespread use, which remains a challenge primarily due to the greater heterogeneity in scanning protocols of \textit{ex vivo} MRI and increased complexity and imaging artifacts than \textit{in vivo} MRI. Recent developments for \textit{ex vivo} MRI include deep learning methods for high resolution cytoarchitectonic mapping in 2D histology \cite{amunts}, atlas-based segmentation of MTL and the thalamus \cite{Iglesias_hippo}, \cite{Iglesias_thalamus}, and whole-brain hemispheres analysis \cite{rush_exvivo}, \cite{jonkman}. Recently, deep learning-based methods were developed to segment the granular cortical layers \cite{fischl_supragranular}, and gray matter (GM), subcortical structures, white matter (WM) and its hyperintensities (WMH) in 7T \textit{ex vivo} MRI. NextBrain \cite{nextbrain} provides semi-automatic brain segmentations using histology brain atlas.

These previous methods have several limitations as they are specialized for specific parts of the brain or specific \textit{ex vivo} datasets mainly limited to a healthy (unremarkable) brain tissue specimens. The methods were evaluated on very small datasets of only 1 \cite{edlow}, 5 \cite{nextbrain}, and 17 \cite{fischl_supragranular} \textit{ex vivo} brain tissue for unremarkable specimens or at low-resolution \cite{rush_exvivo}. Crucially, to our knowledge, \textbf{the feasibility of using existing \textit{ex vivo} MRI analysis pipelines to perform large-scale structure-pathology association studies in ultra-high resolution \textit{ex vivo} MRI has not been demonstrated}.

\textbf{Contributions.} In this study, we perform structure-pathology association analysis in a large dataset of ultra high-resolution 0.3 mm$^3$ T2w 7T MRI scans of whole brain hemispheres from 82 brain donors with ADRD diagnoses, a first study of such scale conducted at this resolution. We present a new computational pipeline that performs automated segmentation and \textbf{whole-hemisphere FreeSurfer DKT atlas \cite{Desikan} parcellation of the cortex in native subject space} at sub-millimeter 0.3 mm $^3$ resolution, a \textbf{first large-scale surface-based scheme for \textit{ex vivo} whole-hemispheres analysis in diseased population}. We achieve this by adapting the surface-based modeling steps in FreeSurfer with an initial subject-space topology-corrected WM segmentation derived from a deep learning-based segmentation model as developed in \cite{Khandelwal}. We evaluate the framework by correlating cortical thickness with neuropathological markers implicated in AD (measures of p-tau, neuronal loss; global amyloid-$\beta$, Braak staging, and CERAD ratings) and perform vertex-wise generalized linear modeling.

\section{Materials and Methods}
\label{sec:Methodology}
\vspace{-3mm}
\hspace{5 mm}\textbf{Dataset.}
\label{sec:Dataset}
We acquire and analyze a dataset of \textbf{82 ex vivo whole hemisphere 0.3 mm$^3$ 7T MRI} (See Fig. \ref{fig:2}) at our XYZ center with left: 42 and right: 40 hemispheres, postmortem interval (PMI) of 18.48 $\pm$ 13.60 hours and fixation time of 256.70 $\pm$ 280.48 days. The cohort included patients with Alzheimer’s Disease or related dementias (ADRD) \cite{jack} spectrum, such as Lewy body disease, limbic-predominant age-related TDP-43 encephalopathy, corticobasal degeneration, primary age-related tauopathy comprising of 41 female (age: 76.97 $\pm$ 9.70 years) and 41 male (age: 76.48 $\pm$ 11.67 years). Human brain specimens were obtained in accordance with local laws and regulations, and includes informed consent from next of kin at time of death. After autopsy, one hemisphere (Supplemental Fig. 1) was fixed in formalin for atleast 4 weeks and then imaged at T2-w 7T using a 3D-encoded T2 SPACE sequence with 0.3 mm$^3$ isotropic resolution, 3s repetition time (TR), echo time (TE) 383 ms, turbo factor 188, echo train duration 951 ms, bandwidth 348 Hz/px with 2-3 hours per scan.

The non-imaged hemisphere (contralateral tissue) underwent histological processing for neuropathological examination (Supp. Fig. 2). Tissue blocks were embedded in paraffin, sectioned at 6 $\mu$m thickness, and underwent immunohistochemistry. In each region, semi-quantitative severity ratings of p-tau pathology, amyloid-$\beta$, and neuronal loss, were assigned by expert neuropathologists on a scale of 0–3 i.e. `0: None/Rare', `1: Mild', `2: Moderate' or `3: Severe'. Standardized global AD progression ratings were derived, including Thal stage (global amyloid-$\beta$), Braak (global p-tau), and CERAD stage (neuritic plaques) \cite{cerad}.

\begin{figure}[t]
\centering
\includegraphics[width=\textwidth,height=0.30\textheight]{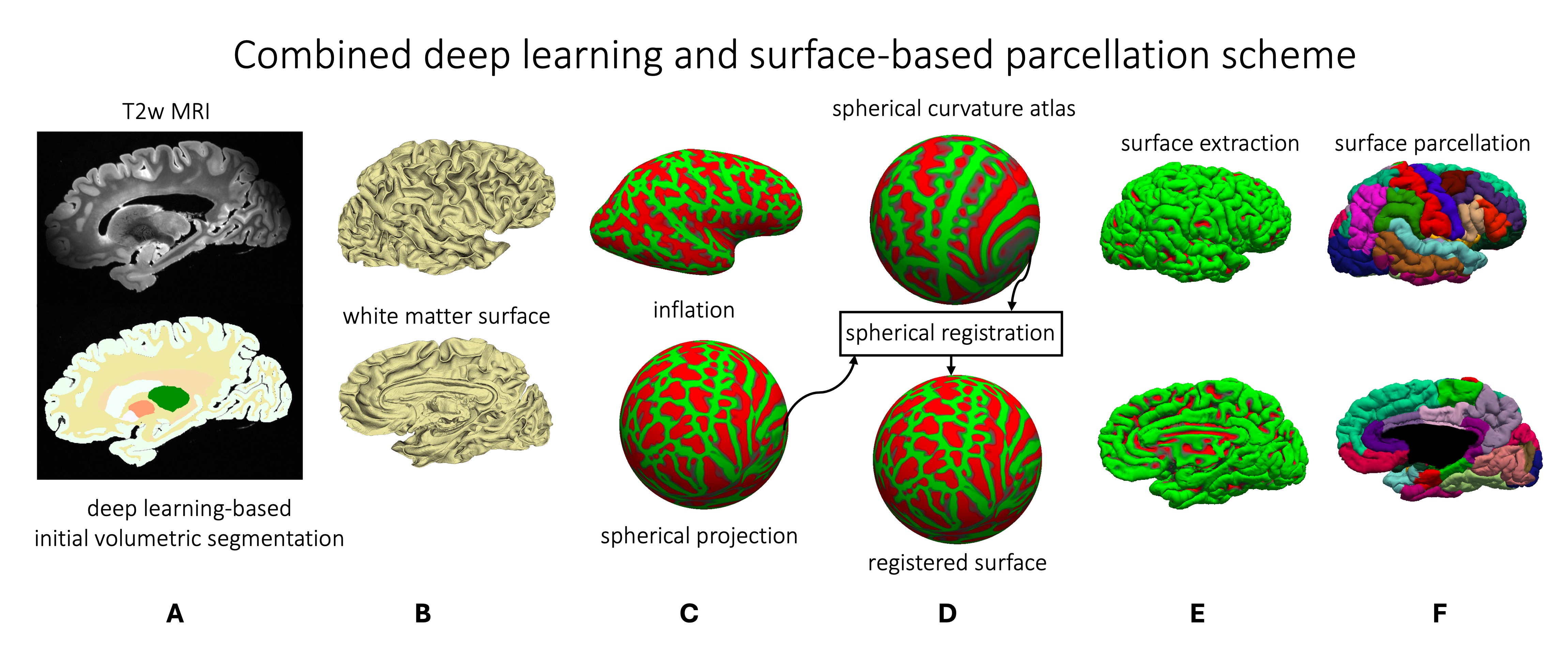}
        \caption{Schematic of the developed pipeline based on deep learning volumetric segmentations and surface-based modeling for parcellations of \textbf{\textit{ex vivo} whole hemisphere 0.3 mm$^3$ 7T MRI}. The sequential steps follow A-F as described in Section 2.}
        \label{fig:1}
\end{figure}

\begin{figure}[h!]
\centering
\includegraphics[width=0.95\textwidth,height=0.45\textheight]{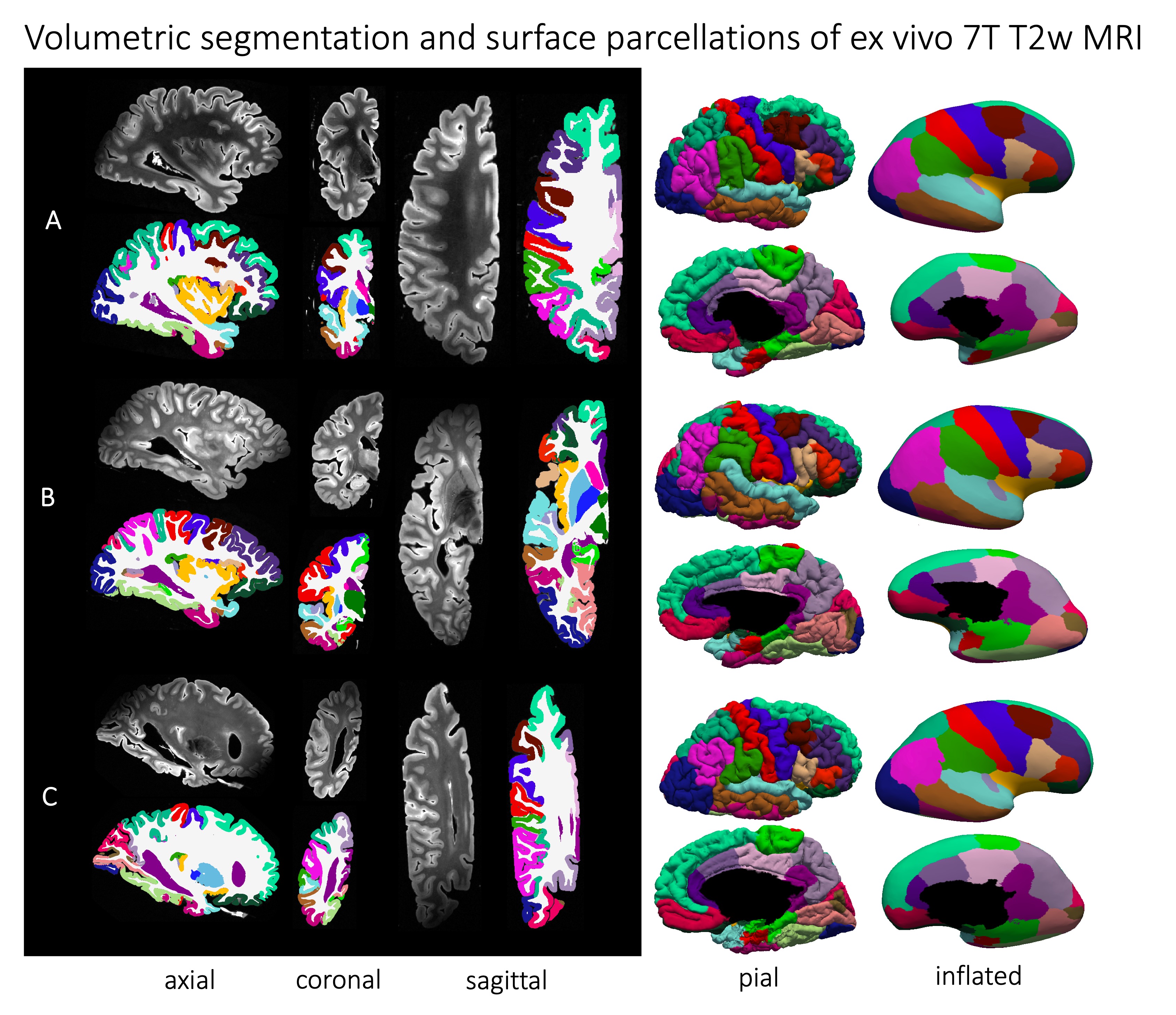}
    \caption{\textbf{Ex vivo MRI segmentations and parcellations.} Axial, coronal and sagittal viewing planes of \textbf{\textit{ex vivo}} MRI at 0.3 mm$^3$ resolution for three subjects (A, B and C) with corresponding DKT volumetric segmentations and surface-based parcellations on pial and inflated surfaces for the medial and lateral views in native subject space resolution. Our method is able to correctly delineate the brain even in regions where the MR signal contrast is low in the anterior and the posterior brain MRI due to artifacts in acquisition protocol. Legend: See Fig. \ref{fig:3}.}
    \label{fig:2}
\end{figure}

\textbf{Methodological pipeline.}
\label{sec:Pipeline}
We present a fast and reliable anatomical parcellation and cortical thickness quantification framework that combines deep learning-based methods developed in \cite{Khandelwal} with classical methods for topology correction \cite{Segonne} and surface-based cortical modeling, inflation, registration, and parcellation in FreeSurfer [\cite{Fischl_one}, \cite{Fischl_two}] for whole-hemisphere \textbf{\textit{ex vivo}} brain MRI. Our framework processes one hemisphere in around 60 minutes. Rather than reinventing the wheel, our approach focuses on combining the strengths of these existing tools and making modifications that enable them to scale to ultra-high resolution \textbf{\textit{ex vivo}} MRI. See Fig. \ref{fig:1} for the schematic pipeline.

\textbf{Volume-based segmentation:} First, we train a deep learning segmentation model nnU-Net \cite{nnunet} as explained in \cite{Khandelwal} to obtain whole-hemisphere volume-based segmentations of 8 regions namely: cortical gray matter (GM), white matter (WM), white matter hyperintensities (WMH), ventricles and four subcortical structures: caudate, putamen, globus pallidus and thalamus (Fig. \ref{fig:1} column A). The model was trained on Nvidia Quadro RTX 5000 GPU using manual-labeled images and extensively validated, as detailed in the work by Khandelwal et al. \cite{Khandelwal}. The inference time for each hemisphere is around 15 minutes on a CPU. We merge the WM, WMH, ventricles and subcortical labels in a single label and employ the post-hoc topology correction method [\cite{CRUISE} and \cite{bazin}] based on fast-marching algorithm implemented in `Nighres/CRUISE' \cite{Nighres} to solve the \textit{buried sulcus} problem, i.e., clearly separating the adjoining gyri in the opposite banks of a buried sulcus. This correction step takes around 5 minutes on a CPU.

\textbf{Surface-based parcellation:} Next, we modify the surface-based scheme in FreeSurfer to obtain DKT atlas-based surface parcellations in \textit{ex vivo} native subject space. Of note is that this step does not require the raw MRI intensity image and is purely based on the voxel-level segmentation of cortical GM from above. First, a `filled' segmentation with WM, WMH, ventricles and the four subcortical structures as the foreground; and the cortical GM along with the CSF as the background, is created from the initial volume-based segmentation. This `filled' segmentation (Fig. \ref{fig:1} column B) is then tessellated and corrected for any topological errors such as holes and handles based on \cite{Segonne}, termed as the `WM surface'. The `WM surface' is smoothed and inflated to a sphere (Fig. \ref{fig:1} column C) which is then registered to the spherical representation of the Buckner40 spherical atlas \cite{buckner} provided in FreeSurfer using the curvature map of the WM surface (Fig. \ref{fig:1} column D). The labels of the DKT-atlas is then warped from the spherical-atlas to the spherical-representation of the \textit{ex vivo} subject. Next, the `WM surface' is deformed and re-positioned to find the GM-CSF pial surface using the method described in \cite{Fischl_one}\, \cite{Fischl_two} which employs three energy functionals: spring-like term, decomposed into tangential and normal components, to smooth and for regularization of the surface. The third term in the energy functional is dependent on the gradient of the intensity image. This term is the classic stopping criteria (inverse of the image gradient) used in deformable models based on level sets and fast marching methods. FreeSurfer uses raw \textit{in vivo} T1w (or where applicable T2w) MRI intensity values for this energy term. For the current study, the T2w \textit{ex vivo} MRI has a varying intensity profile within the GM which causes the deformable surface to prematurely stop in the GM cortical ribbon and does not reach the pial surface. Therefore, we use the segmentation derived from the above volume-based step in place of the T2w \textit{ex vivo} MRI intensity image. This has a clear advantage as the intensity is constant in the GM cortical ribbon with a zero gradient everywhere. The evolving surface then stops at the pial surface as expected (Fig. \ref{fig:1} column E). Finally, the DKT-atlas labels are projected onto the pial surface and the parcellations (Fig. \ref{fig:1} column F) are mapped to the voxel-space to get the region-wise segmentation in the volume. This surface-based scheme takes 40 minutes on a CPU with 64 threads.

\begin{figure}[t]
\centering
\includegraphics[width=\textwidth,height=0.3\textheight]{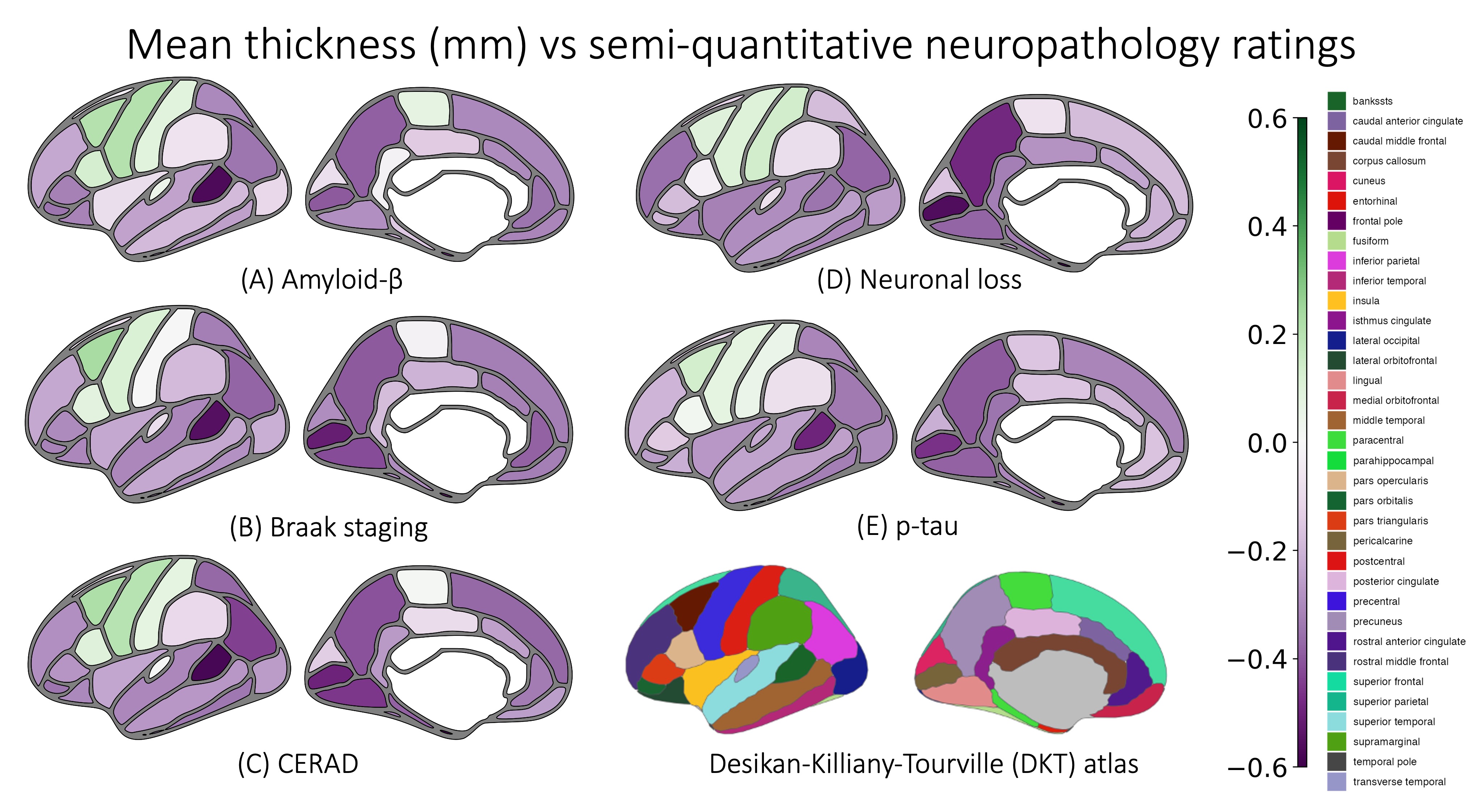}
        \caption{\textbf{Spearman’s correlation plots between mean ROI thickness (mm) and neuropathological ratings in native subject-space.} We observe significant negative correlation with global ratings of amyloid-$\beta$, Braak staging, CERAD, and the semi-quantitative ratings of the medial temporal lobe (MTL) neuronal loss and tau pathology. All the analysis were covaried for age, sex and postmortem interval (PMI) for the entire cohort of 82 subjects. See legend for the regional brain labels.}
        \label{fig:3}
\end{figure}
\vspace{-3mm}
\setlength{\textfloatsep}{10pt plus 1.0pt minus 2.0pt}

\section{Experiments and Results}
\vspace{-3mm}
\hspace{5 mm}\textbf{Segmentation and parcellation.}
\label{sec:segm_parc}
We deployed the deep-learning model to produce the initial segmentation of GM, WM, WMH, ventricles and the subcortical structures. The deep learning-based pipeline was extensively validated in a previous work \cite{Khandelwal} as explained in Sec. \ref{sec:Pipeline}. The surface-based scheme was then used to jointly obtain the cortical parcellations and volumetric segmentations as shown in Fig. \ref{fig:2} in the native subject space. Due to the immense manual labelling efforts involved in\textit{ ex vivo} MRI, in the order of 3-4 weeks per subject, the reliability of these parcellations are evaluated using the following two correlation studies.

\textbf{Region-based cortical thickness vs neuropathology correlations.}
\label{sec:corr}
Fig. \ref{fig:3} shows the Spearman's $\rho$-value between the mean cortical thickness (mm) in each brain region (computed in subject space at native resolution) and five pathology measures: global ratings of amyloid-$\beta$, Braak staging, CERAD, and regional ratings of neuronal loss and p-tau pathology in the MTL, the region first implicated in AD. The analysis were covaried for age, sex and postmortem interval (PMI) and corrected for multiple comparisons using Bonferroni method. Significant negative correlation were found in entorhinal, parahippocampal, medial-orbitofrontal, temporal pole, inferior temporal and parietal lobes which are consistent with literature on progressive loss of cortical gray matter in AD \cite{sadaghiani}.

\begin{figure}[h]
\centering
\includegraphics[width=\textwidth,keepaspectratio]{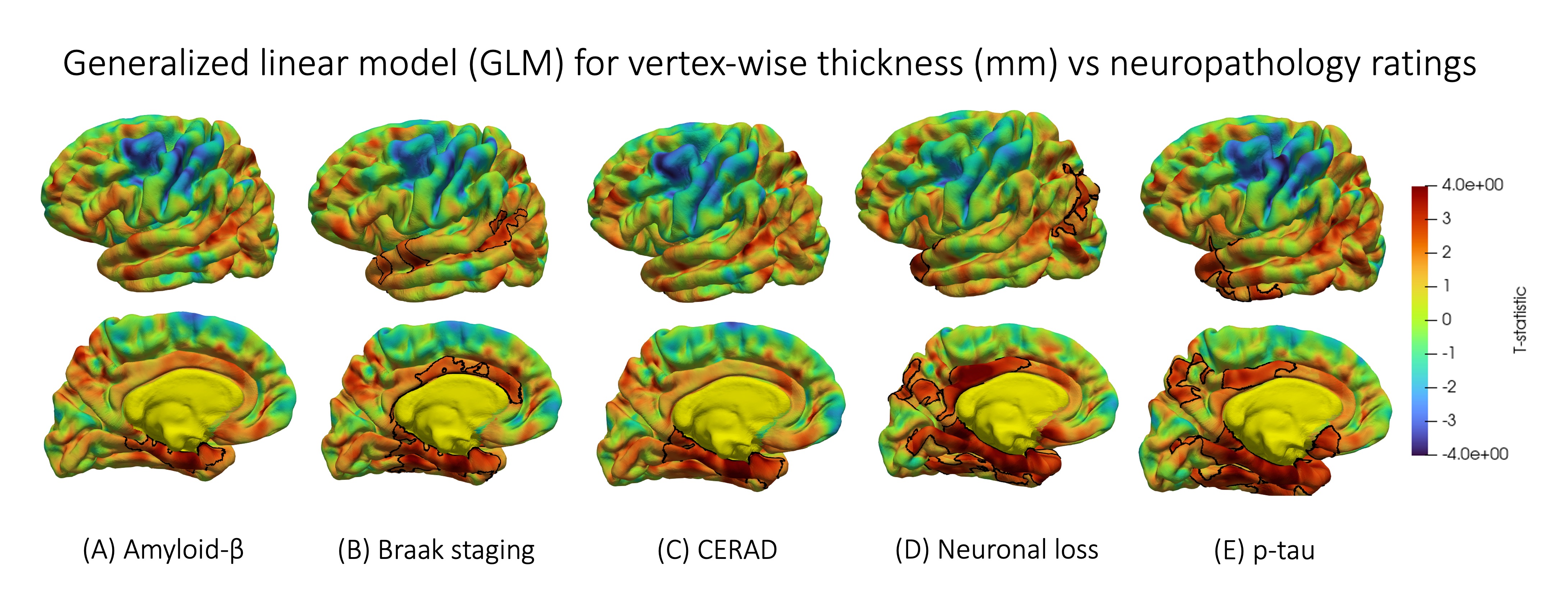}
        \caption{\textbf{Template-space vertex-wise morphometry-pathology correlations.} Vertex-wise group analysis was performed to fit a generalized linear model (GLM). Shown are the statistical map (t-statistics) of the correlation between cortical thickness (mm) and with global ratings of amyloid-$\beta$, Braak staging, CERAD, and semi-quantitative ratings of the medial temporal lobe (MTL) neuronal loss and tau pathology, with age, sex and postmortem interval (PMI) as covariates across all 82 subjects. The clusters outlined in black indicate regions significant correlations (p$<$0.05) were observed after FWER correction for multiple comparisons.}
\label{fig:4}
\end{figure}

\textbf{Surface-based vertex-wise cortical thickness vs neuropathology correlations.}
\label{sec:group_analysis}
The thickness maps for each individual subject were warped to the Buckner40 template-space for vertex-wise correlation analysis between cortical thickness (mm) and the five neuropathological ratings. The parcellations and thickness maps of the three subjects (Fig. \ref{fig:1}) in the Buckner40 \textit{fsaverage} space are shown in Supplemental Fig. 3. Vertex-wise correlation between thickness and the neuropathology ratings was performed by fitting a GLM at each vertex across the entire cohort of 82 subjects with age, sex and PMI as the covariates and corrected for multiple comparisons using family-wise error rate (FWER) correction. Fig. \ref{fig:4} shows the t-statistics map on the pial and the inflated surfaces with the clusters outlined in white indicating regions where the significant strongest associations were observed (p$<$0.05) surviving FDR correction. We observe that the strongest correlations were observed in MTL, the region associated with AD.

\textbf{Comparison with other methods and parcellations with other atlases.}
\label{sec:other_methods}
We compared our method with two other segmentation methods SynthSeg \cite{synthseg} (designed for contrast-agnostic MRI) and NextBrain \cite{nextbrain} (designed specifically for ex vivo MRI). See Supplemental Fig. 4 for the qualitative results. Both SynthSeg and NextBrain fail to provide meaningful segmentations of \textit{ex vivo} MRI. SynthSeg (operating at lower 1 mm$^{3}$) roughly delineates major brain regions but fails at segmenting the finer details in the cortical ribbon and bleeds into CSF. Whereas, NextBrain incorrectly segments the entire cortical ribbon as it does not demarcate the different sulci and gyri. In contrast, our method clearly demarcates the brain regions in native 0.3 mm$^{3}$ subject-space high resolution using the DKT-atlas. After visual quality control of segmentations produced by both SynthSeg and NextBrain across all the 82 ex vivo subjects, we concluded that a quantitative analysis similar to our proposed method and experiments is not feasible as depicted in Supp. Fig. 4.
Separately, in addition to the DKT atlas, we also provide parcellations using the Schaefer \cite{schaefer}, Glasser \cite{glasser} and the Von Economo-Koskinos \cite{economo} atlases in native subject-space resolutions for \textit{\textbf{ex vivo}} brain hemisphere as shown in Supplemental Fig. 5.

\setlength{\textfloatsep}{10pt plus 1.0pt minus 2.0pt}
\vspace{-1mm}
\section{Discussion}
\vspace{-5mm}
\hspace{5 mm}Using both recent advances in deep learning based segmentation models and classic deformable and spherical registration methods, we developed a pipeline that enables surface-based modeling and group-wise registration of the cortical gray matter in ultra high-resolution \textit{ex vivo} MRI. In addition to its higher resolution, \textit{ex vivo} MRI poses multiple challenges for computational analysis, including deformation during brain removal and fixation (particularly collapse of the ventricles), and distinct artifacts, such as intensity inhomogeneity both across the whole image domain and across cortical layers. Hence our ability to successfully combine nnU-Net, Nighres/CRUISE and FreeSurfer pipelines to achieve the same type of parcellation as is ubiquitous in \textit{in vivo} MRI studies is a non-trivial accomplishment. Crucially, the Nighres/CRUISE and FreeSurfer components used in our pipeline operate only on segmentation maps, rather than MRI intensity. Thus, to generalize our approach to other scanners and ultra-high resolution MRI protocols, one only needs to focus on the segmentation aspect.

A key contribution of this work is that it performs, for the first time, a surface-based structure-pathology correlation analysis using ultra-high resolution \textit{ex vivo} MRI in a large cohort of brain donors with ADRD diagnoses. By doing so, it demonstrates the analysis approach that is ubiquitous in \textit{in vivo} brain MRI research, and has been previously adopted to  $1\mathrm{mm}^3$ resolution \textit{ex vivo} MRI by some groups \cite{rush_exvivo} and \cite{jonkman}, can also be applied to \textit{ex vivo} studies at ultra-high resolution. To demonstrate analysis modes common in \textit{in vivo} research, we performed both region of interest (ROI)-level correlation analysis and vertex-wise template-space analysis to study patterns of neurodegeneration within the AD continuum. Prior works [\cite{cerad_corr}, \cite{Frigerio}] shows how amyloid-$\beta$ and p-tau etc have differential influences on cortical atrophy. Tau pathology is concurrent with neuronal loss in AD leading to cortical atrophy \cite{Harrison}. Negative correlations between MTL p-tau and neuronal loss with cortical thickness were found to be significant in the entorhinal cortex and the MTL, regions first implicated in AD [\cite{das}, \cite{Renaud}] both in \textit{in vivo} and \textit{ex vivo} studies. Crucially, the ability to map data from subject space to the DKT-template space based on surface correspondence is not limited to cortical thickness measures, and in future, our pipeline can be used to analyze features that take advantage of ultra-high resolution \textit{ex vivo} MRI, such as mapping iron and myelin in cortex \cite{Tisdall}.

\textbf{Conclusion.} We developed a fast and easy-to-use pipeline to parcellate the whole-brain hemispheres into the FreeSurfer DKT-atlas defined regions at the native subject space sub-millimeter 0.3 mm$^3$ isotropic resolution, and applied it to a  large-scale dataset of ultra high-resolution \textit{ex vivo} 7T structural MRI.  We demonstrated the feasibility and utility of this approach by performing a structure-pathology association study that is the first of its kind at this level of \textit{ex vivo} MRI resolution and one of the largest ever conducted at any \textit{ex vivo} MRI resolution. In future, we will perform an exhaustive quantitative study by delineating the brain regions with different anatomical atlases based on a surface-based population template derived from the \textit{ex vivo} hemispheres. This will enable us to understand the region-wise local intensity profiles to study the structural relationships with the underlying histology-derived markers. We believe that the present study will enable scientific advancements in \textit{ex vivo} imaging and thereby inform better \textit{in vivo} computational biomarkers.

\counterwithin{figure}{section}
\appendix
\section{Supplementary Figures}
\begin{figure}
\centering
\includegraphics[width=\textwidth,height=\textheight,keepaspectratio]{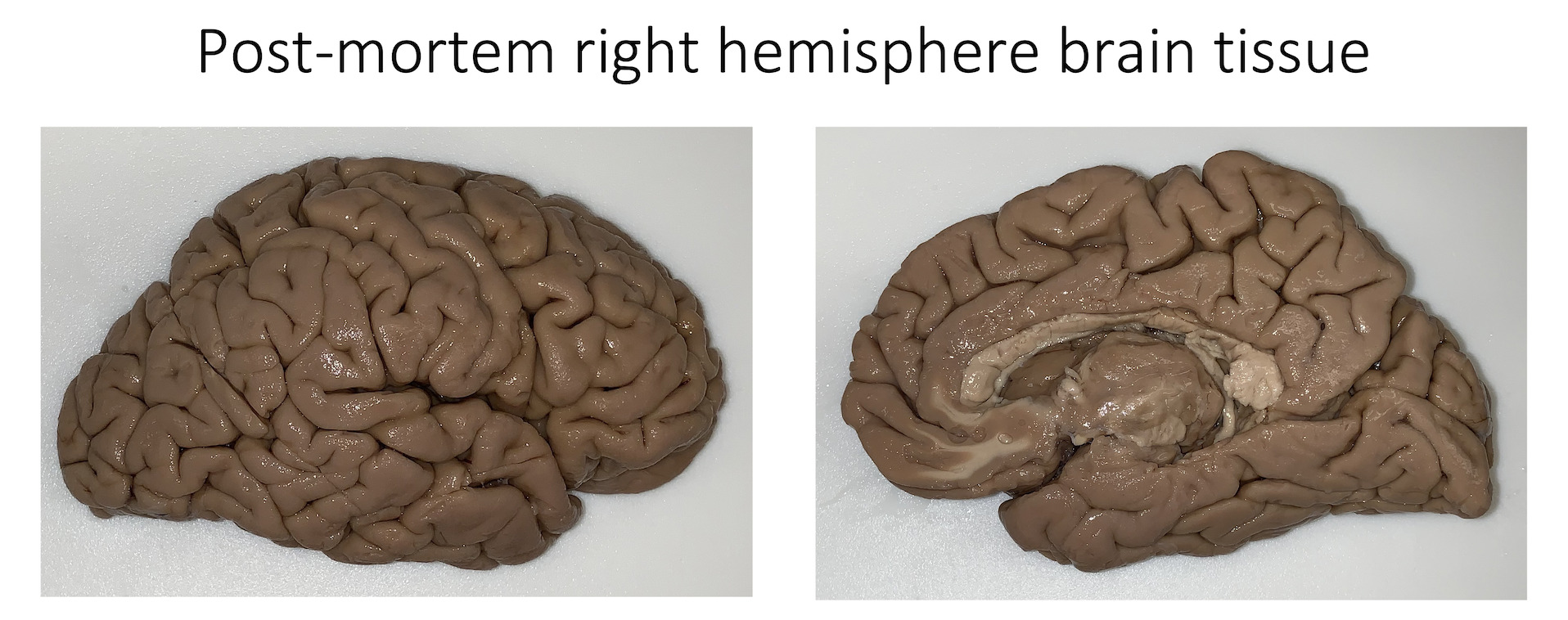}
\caption{Ex vivo tissue blockface photograph of a donor with co-morbid diagnosis of Parkinson’s disease and Lewy body disease (deceased at the age of 79). Shown are the lateral and the medial views of the right hemisphere.}
\end{figure}

\begin{figure}
\centering
\includegraphics[width=\textwidth,height=\textheight,keepaspectratio]{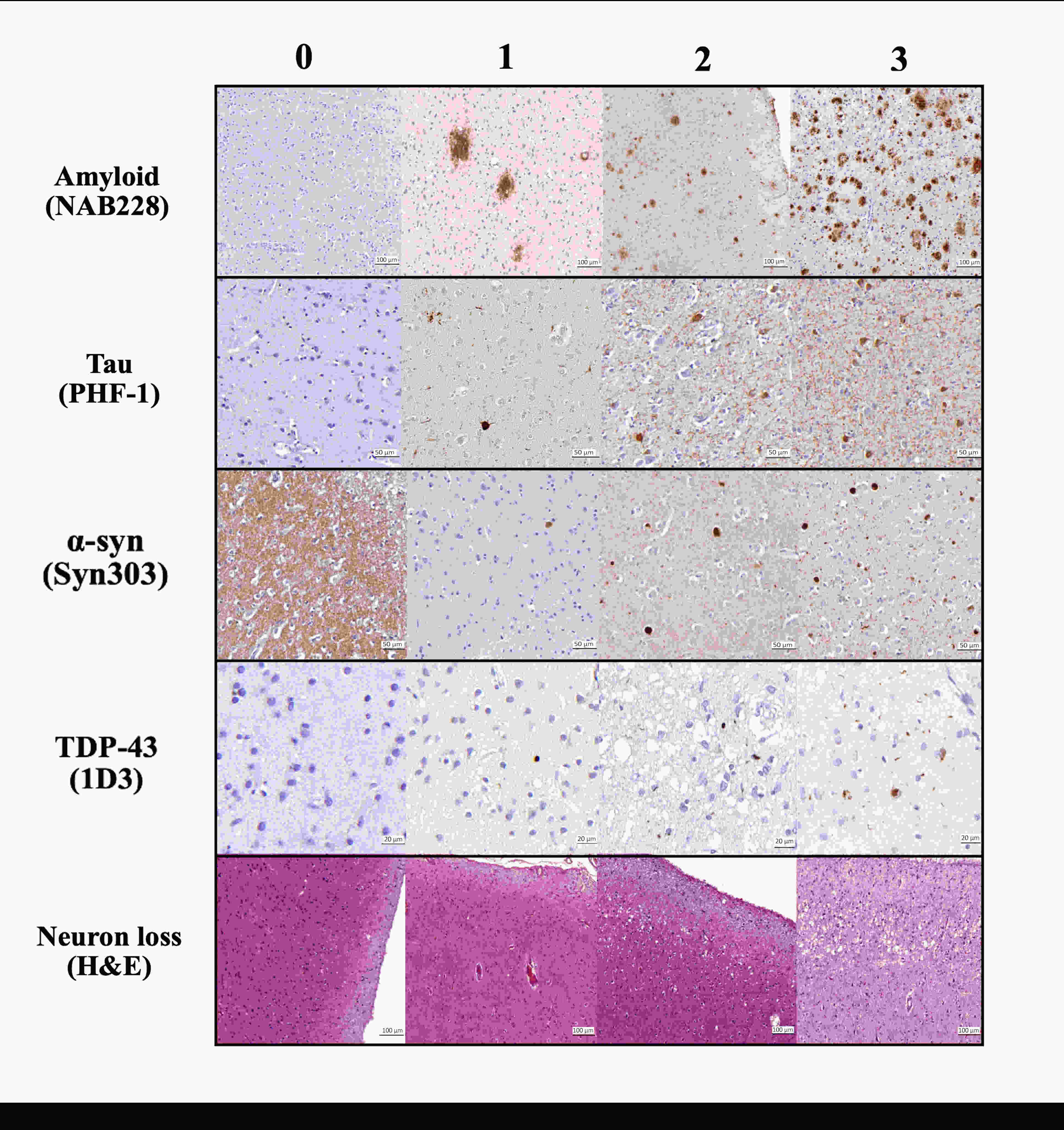}
\caption{ Pathological assessment for each regional pathology derived from histology. The columns are severity ratings (left to
right): 0-3. The shown pathologies are: amyloid-$\beta$ plaques (NAB228 antibody), p-tau pathology (PHF-1 antibody), and neuron loss (Hematoxylin and Eosin staining).}
\end{figure}

\begin{figure}
\centering
\includegraphics[width=\textwidth,height=1.4\textheight,keepaspectratio]{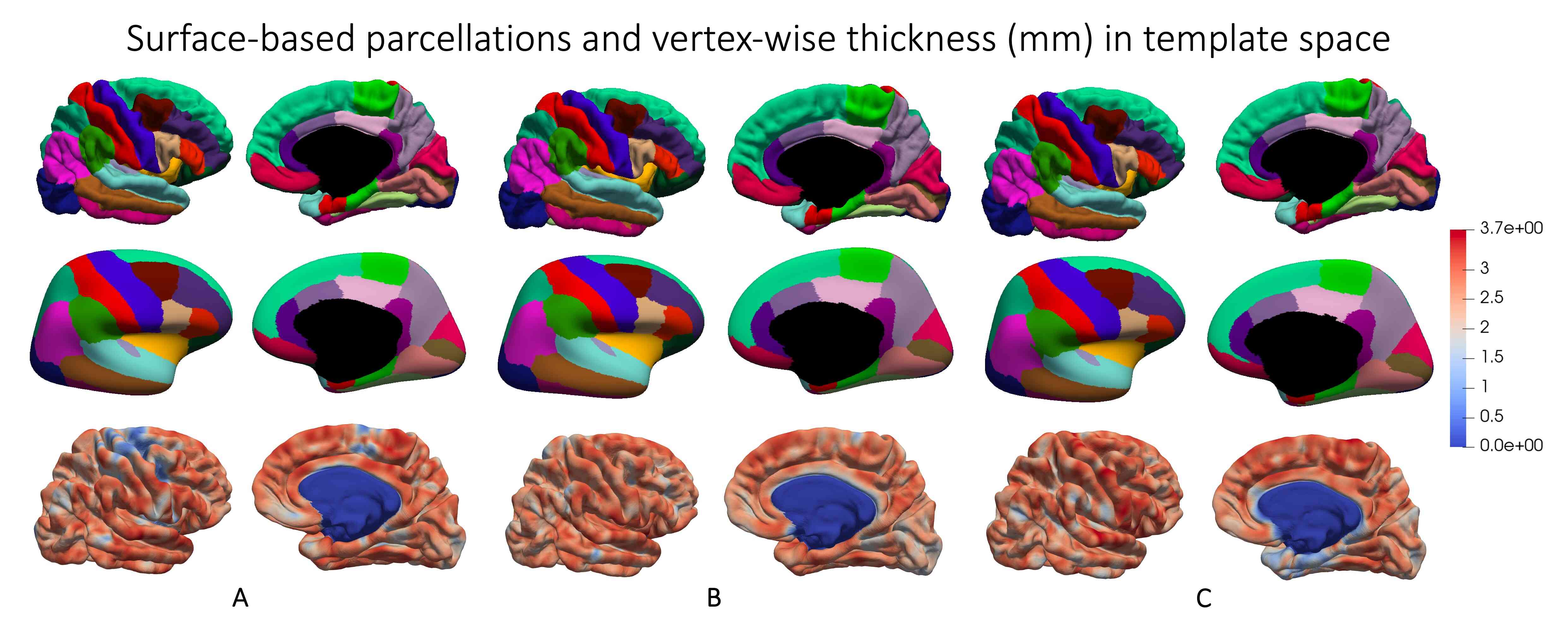}
\caption{\textbf{Desikan-Killiany-Tourville (DKT) surface-based parcellations and thickness maps} shown for the three ex vivo subjects (A, B and C) in Main Fig. 2 in the FreeSurfer Buckner40 template-space on pial and inflated surfaces for the medial and the lateral views. The legend for the DKT atlas can be found in Fig. 2 in the main manuscript. The thickness maps are smoothed with a 5mm FWHM Gaussian kernel and are used for the vertex-wise GLM analysis as described in the main paper.}
\end{figure}

\begin{figure}
\begin{subfigure}{\textwidth}
  \centering
  \includegraphics[width=0.96\textwidth,height=\textheight,keepaspectratio]{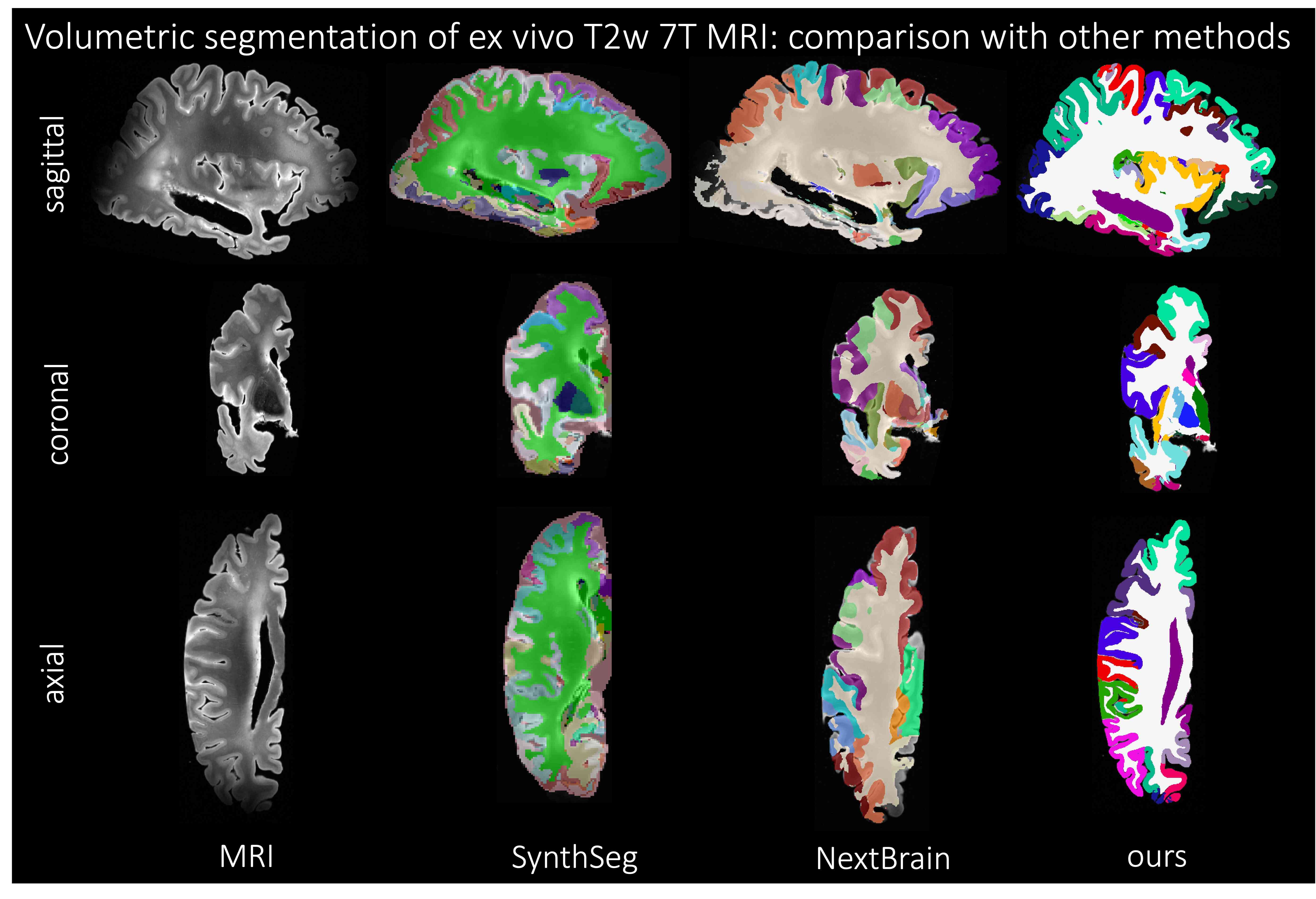}
\end{subfigure}%

\begin{subfigure}{\textwidth}
  \centering
  \includegraphics[width=0.95\textwidth,height=\textheight,keepaspectratio]{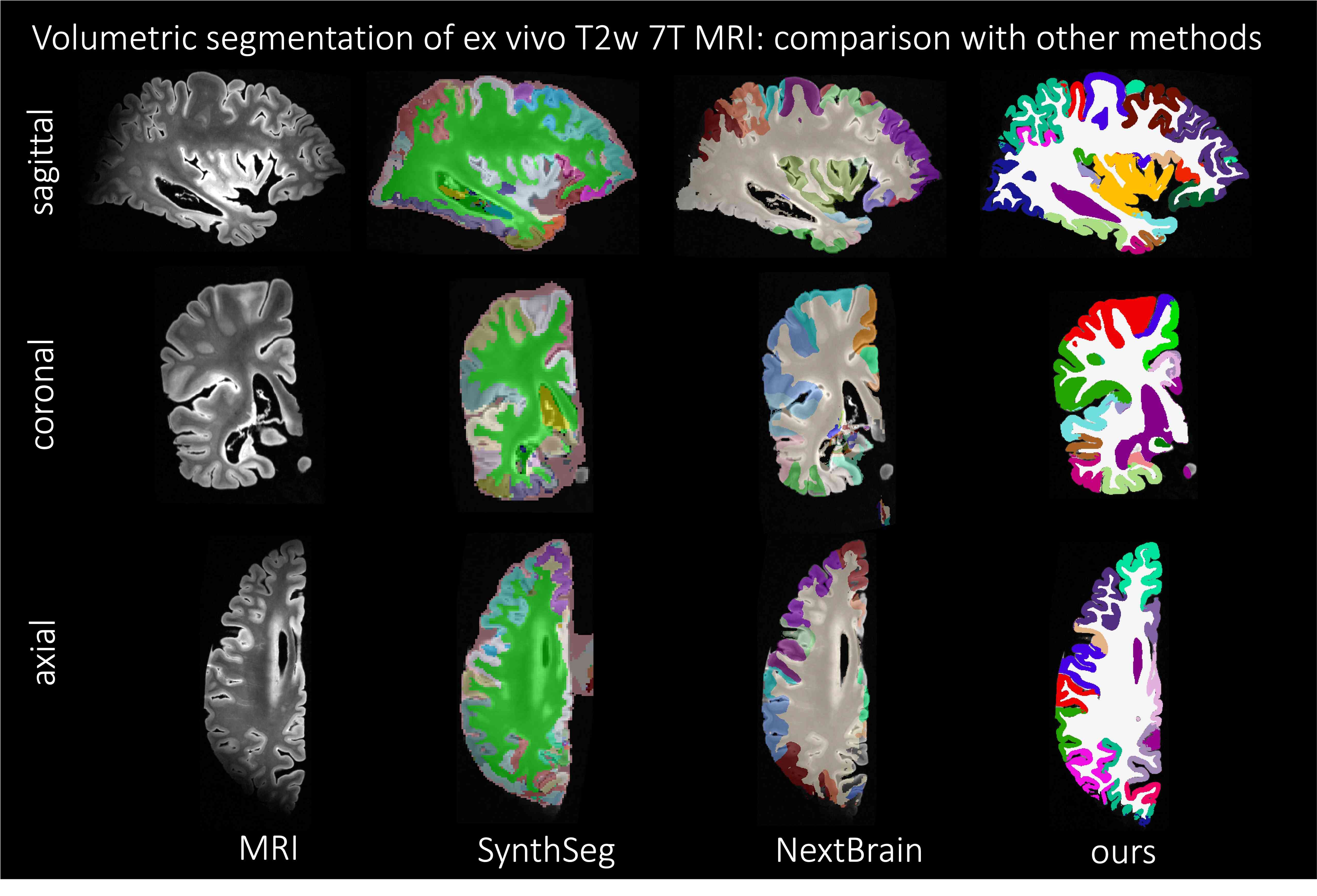}
\end{subfigure}
\caption{Comparative study of different methods for ex vivo MRI segmentation. We compared our method with two other segmentation methods SynthSeg \cite{synthseg} and NextBrain \cite{nextbrain}. Shown are the qualitative results for two subjects (1) and (2) in the three viewing planes. Both SynthSeg and NextBrain fail to provide meaningful segmentations of the ex vivo brain hemisphere. SynthSeg roughly delineates the major brain regions but fails at segmenting the finer details in the cortical ribbon. SynthSeg produces the segmentation by up-sampling to 1 mm$^{3}$ lower-resolution which loses the strength of the sub-millimeter high-resolution imaging. NextBrain incorrectly segments the entire cortical ribbon as it does not demarcate the different sulci and gyri. In contrast, our method clearly demarcates the brain regions in the native 0.3 mm$^{3}$ subject-space high resolution using the DKT-atlas. After visual quality control of segmentations produced by both SynthSeg and NextBrain across all the 82 ex vivo subjects, we concluded that a quantitative analysis similar to our proposed method and experiments is not feasible.}
\label{fig:fig}
\end{figure}

\begin{figure}
\centering
\includegraphics[width=\textwidth,height=\textheight,keepaspectratio]{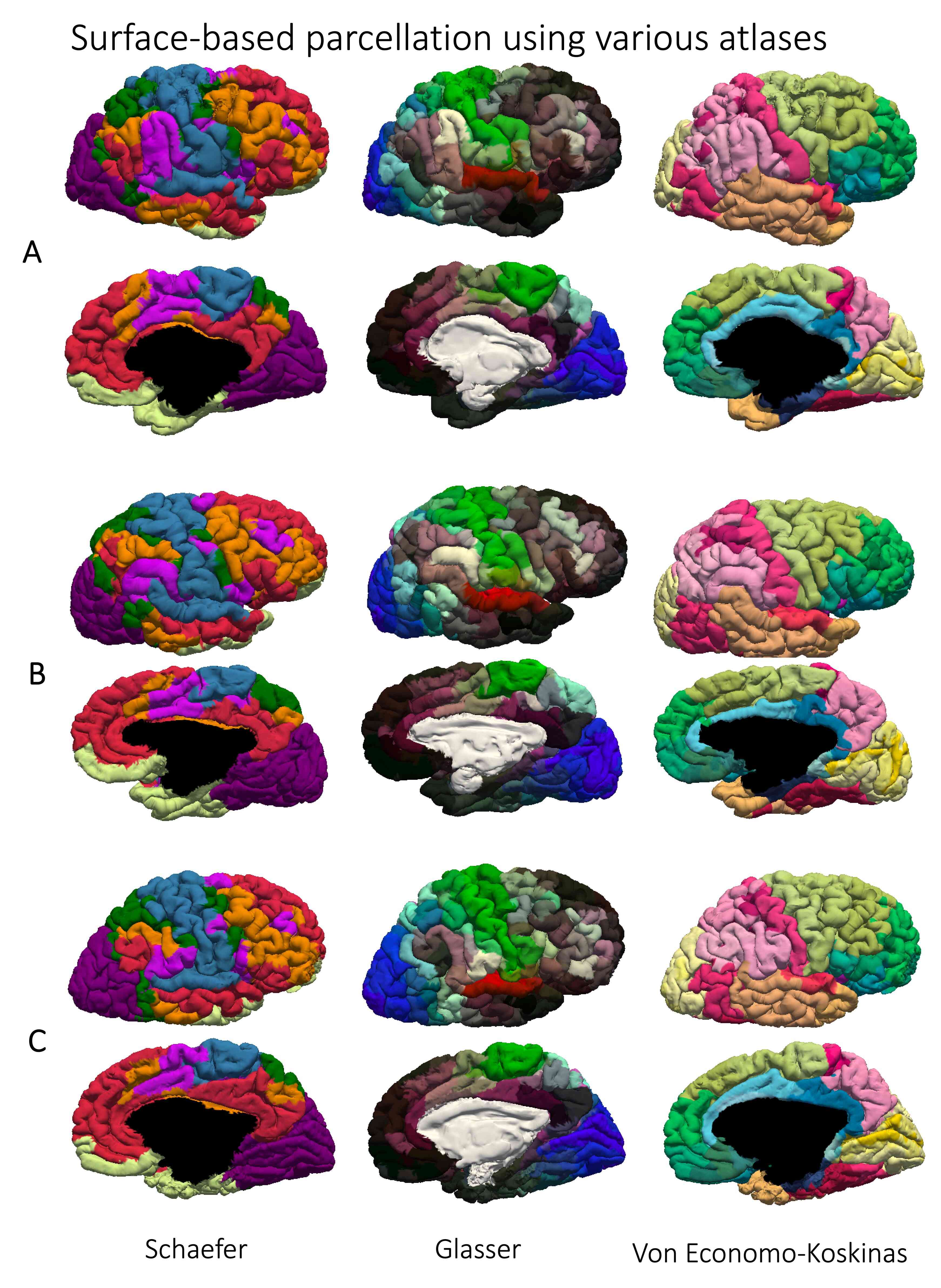}
\caption{\textbf{Schaefer, Glasser and the Von Economo-Koskinos surface-based parcellations for ex vivo MRI.} Shown here are the surface-based parcellations on pial and inflated surfaces for the medial and lateral views for the three subjects (A, B and C) from Fig. 2 in the main paper in native subject-space resolution for the Schaefer \cite{schaefer}, Glasser \cite{glasser} and the Von Economo-Koskinos \cite{economo} atlases.}
\end{figure}

\end{document}